\documentclass[10pt,a4paper,twocolumn]{article}

\usepackage{mathtools}
\usepackage{amsmath,amsthm,amssymb}
\usepackage{bm}

\usepackage[T1]{fontenc}
\usepackage{newtxtext}          
\usepackage[cmintegrals]{newtxmath}  
\usepackage{microtype}
\usepackage[
  a4paper,
  top=2cm,
  bottom=2cm,
  left=1.8cm,
  right=1.8cm,
  columnsep=0.6cm,
  headheight=14pt
]{geometry}

\usepackage{graphicx}
\usepackage{float}
\usepackage{booktabs}
\usepackage{multirow}
\usepackage{tabularx}
\usepackage{makecell}           
\usepackage{caption}
\usepackage{subcaption}

\usepackage[dvipsnames]{xcolor}
\definecolor{secgray}{HTML}{333333}

\definecolor{linkdark}{HTML}{1A1A2E}     
\definecolor{citedark}{HTML}{2E4057}     
\definecolor{urlteal}{HTML}{3A6B7E}      

\usepackage[
  colorlinks=true,
  linkcolor=linkdark,
  citecolor=citedark,
  urlcolor=urlteal,
  bookmarks=true,
  bookmarksnumbered=true,
  pdfstartview={FitH},
]{hyperref}
\usepackage[nameinlink,capitalize,noabbrev]{cleveref}

\usepackage[numbers,sort&compress,super]{natbib}
\usepackage{titlesec}

\titleformat{\section}
  {\large\bfseries\color{secgray}}
  {\thesection}{0.6em}{}
\titlespacing*{\section}{0pt}{14pt plus 3pt minus 2pt}{4pt plus 2pt minus 1pt}

\titleformat{\subsection}
  {\normalsize\bfseries\color{secgray}}
  {\thesubsection}{0.6em}{}
\titlespacing*{\subsection}{0pt}{10pt plus 2pt minus 1pt}{2pt plus 1pt minus 1pt}

\titleformat{\paragraph}[runin]
  {\normalsize\bfseries}
  {}{0em}{}[.]
\titlespacing*{\paragraph}{0pt}{6pt plus 2pt minus 1pt}{0.5em}

\usepackage{fancyhdr}
\fancypagestyle{plain}{%
  \fancyhf{}
  \fancyfoot[C]{\small\thepage}
  
}

\usepackage{lineno}

\usepackage[title]{appendix}

\newtheorem{theorem}{Theorem}

\newtheorem{corollary}{Corollary}

\DeclareMathOperator{\Tr}{Tr}
\usepackage{braket}             

\newcommand{\dsys}{d_{\text{sys}}}
\newcommand{\denv}{d_{\text{env}}}
\newcommand{\dtot}{d_{\text{total}}}

\graphicspath{
  {figures/}
  {../figures/main_text/}
  {../figures/supplement/}
}

\usepackage{enumitem}
\usepackage{authblk}
\usepackage{placeins}
\usepackage{flushend}

\newcolumntype{C}{>{\centering\arraybackslash}X}
\newcolumntype{R}{>{\raggedleft\arraybackslash}X}

\newcommand{\sys}{\mathcal{S}}
\newcommand{\env}{\mathcal{E}}
\newcommand{\tot}{\mathcal{T}}

\begin{document}

\twocolumn[
\begin{center}
  {\LARGE\bfseries UWM-JEPA: Predictive World Models\\[4pt]
   That Imagine in Belief Space\par}
  \vspace{6pt}
  {\normalsize\itshape Unitary density-matrix latent prediction for partially observable JEPA world models\par}
  \vspace{12pt}
  {\large
    \href{https://www.linkedin.com/in/santoshkumarradha}{Santosh Kumar Radha}\textsuperscript{1,*}
    \quad
    \href{https://www.linkedin.com/in/oktay-goktas-8b2881167/}{Oktay Goktas}\textsuperscript{1}
  \par}
  \vspace{6pt}
  {\small\textsuperscript{1}AgentField AI, Toronto, Canada\par}
  \vspace{4pt}
  {\small\textsuperscript{*}Correspondence: \href{mailto:contact@santoshkumarradha.com}{contact@santoshkumarradha.com}, \href{mailto:santosh@agentfield.ai}{santosh@agentfield.ai}\par}
  \vspace{4pt}
  {\small Code: \href{https://github.com/santoshkumarradha/uwm-jepa}{\texttt{github.com/santoshkumarradha/uwm-jepa}}\par}
  \vspace{14pt}
\end{center}

\begin{abstract}
World models for partially observed environments must imagine
multiple compatible hidden futures and steer between them under
counterfactual actions. Joint Embedding Predictive Architectures
(JEPAs) do this in latent space, but a vector-valued latent has no
internal structure for carrying the belief over hidden continuations
through blind rollout. We introduce the Unitary World Model JEPA
(UWM-JEPA), a JEPA world model with a density-matrix latent on a
joint system--environment space and a learned unitary predictor. The
construction preserves the joint-state spectrum exactly during
rollout, so the predictor itself cannot dissipate the represented
uncertainty.

On a hidden-velocity indicator task requiring five-step forward
simulation under a given action sequence with the target observation
masked, UWM-JEPA reaches $0.77$ accuracy and degrades monotonically
as actions are perturbed; a parameter-matched LSTM-JEPA trained
under the same counterfactual-target objective and action head
collapses to majority-class accuracy ($0.53$) under every action
condition. Under blind rollout, UWM-JEPA loses fewer than ten points
of probe $R^2$ at short horizons while vector-latent baselines lose
forty-one and sixty-eight; both nevertheless tie on a held-out
context probe, locating the separation in the predictor rather than
the encoder. Action sensitivity itself requires training against
counterfactual rather than teacher-forced targets, a finding that
applies beyond the unitary parameterisation. For JEPA world models
to imagine under partial observability, latent geometry and
predictor dynamics matter, not frozen context-encoding capacity
alone.
\end{abstract}

\vspace{8pt}
\noindent\rule{\textwidth}{0.4pt}
\vspace{8pt}
]

\section{Introduction}
\label{sec:intro}

JEPA-style world models predict in latent space rather than reconstruct
observations. An online encoder maps an observed context to a latent
representation, a predictor advances that latent forward, and a slowly
moving target encoder supplies the future representation to
match~\cite{grill2020bootstrap,tarvainen2017mean}. The abstraction
underlies I-JEPA~\cite{assran2023ijepa}, V-JEPA~\cite{assran2024vjepa},
and the broader JEPA programme~\cite{lecun2022path}, and it is closely
related to non-contrastive SSL methods such as BYOL~\cite{grill2020bootstrap},
SimSiam~\cite{chen2021exploring}, DINO~\cite{caron2021emerging},
VICReg~\cite{bardes2022vicreg}, and Barlow Twins~\cite{zbontar2021barlow}.
By moving away from pixel- or token-level reconstruction, JEPAs cleanly
separate representation learning from generative decoding, but they
leave open a question that becomes central under partial observability:
what geometry should the latent occupy when it must be imagined forward
without new observations~\cite{ha2018world,hafner2020dream,hafner2020dreamerv2,hafner2023dreamerv3}?

Partial observability changes the object being
predicted~\cite{kaelbling1998planning}. When the current observation
history does not determine the hidden state, the future is not a single
point in representation space. The same context supports several
compatible hidden states, and the natural latent is a belief over those
hidden variables rather than a deterministic
embedding~\cite{kaelbling1998planning,hausknecht2015drqn,igl2018deep,karkus2017qmdp}.
A vector predictor trained against an EMA target can still drive the
loss down, but the loss alone does not specify which uncertainty the
latent should preserve, which should be averaged away, or how a
counterfactual action should move the belief. We study a concrete
alternative: a JEPA whose latent is a density matrix and whose
predictor is a learned unitary on a joint system--environment
space~\cite{nielsen2010quantum,breuer2002theory,stinespring1955positive}.

The resulting model, the Unitary World Model JEPA (UWM-JEPA),
produces a joint density matrix $\rho_t$ from the online encoder and
advances it by
\begin{equation}
  \hat\rho_{t+k\mid t}
  =
  U^k \rho_t (U^\dagger)^k .
\end{equation}
The training scaffold is JEPA-native: an EMA target encoder, a
stop-gradient target, and a latent-space matching loss. Only the
geometry of the predicted quantity changes. A density matrix represents
a graded belief over latent configurations; unitary conjugation
preserves the joint-state spectrum, purity, and von Neumann entropy
exactly during blind rollout. We refer to this property as
\emph{joint-state non-forgetting}.

The qualifier \emph{joint-state} matters. The theorem applies to the
full system--environment density matrix and does not, on its own,
imply that the reduced system state remains linearly decodable after
many steps or that UWM-JEPA encodes context better than a strong
recurrent baseline. The empirical work in this paper measures what
survives the two operations that matter for downstream use: partial
trace to the system state and blind rollout through time.

\paragraph{Counterfactual targets recover an action-sensitive predictor}
Under a teacher-forced JEPA target the learned action term collapses to
$\|H_1\|/\|H_0\|\approx0.03$, because the target encoder has already
consumed the observed future and the predictor can match it without
using the action. Replacing the target by a counterfactual simulator
rollout restores the action term to $1.00\pm0.15$, action perturbations
move the predicted latent in the expected direction, and UWM-JEPA
solves a hidden-velocity indicator task on which the matched
LSTM-JEPA-CF configuration remains at majority-class accuracy.
Additional LSTM-CF and supervised recurrent controls are reported in
the supplement.

\paragraph{The unitary predictor improves short-horizon imagination}
When the latent is rolled forward without new observations, UWM-JEPA
preserves target-nearness better than the vector predictors we tested.
In future-latent retrieval, UWM-JEPA remains far above LSTM-JEPA models
with plain and residual MLP predictors across
$k\in\{1,3,5,10\}$. In teacher-versus-blind probing, UWM-JEPA loses
less than ten points of $R^2$ at the intermediate horizons $k=1$ and
$k=3$, both shorter than the trained horizon $k=5$, while LSTM-JEPA
loses substantially more at those horizons and re-aligns near $k=5$. Beyond $k=10$, both models
have low or negative blind probe scores, so long-horizon retention
remains an open question rather than a settled claim.

\paragraph{Context representation is a control}
On the held-out linear probe used to measure context representation,
UWM-JEPA and a parameter-matched LSTM-JEPA agree within seed noise:
$R^2=0.901\pm0.032$ versus $0.894\pm0.021$ over five seeds (Welch
$p=0.70$). The density-matrix latent is not a stronger context encoder,
so the behavioral separation is not explained by encoder quality.

To our knowledge, UWM-JEPA is the first JEPA-family world model in
which the latent is treated as an explicit belief over hidden
continuations rather than a single representational point, and in
which the predictor is constrained to evolve that belief without
dissipating it during imagined rollout. The density-matrix latent,
the unitary predictor, and the counterfactual action targets are the
specific tools we use to instantiate this property, but the broader
claim is about giving a JEPA latent a place to carry uncertainty and
holding on to that uncertainty under prediction. \Cref{sec:architecture} defines the model.
\Cref{sec:nonforget} states the joint-state guarantee and its scope.
\Cref{sec:actions} gives the main action-conditioned behavioral test,
\cref{sec:imagine} tests blind rollout without new observations, and
\cref{sec:representation} establishes that these effects are not
explained by a stronger context encoder. Collapse diagnostics and the
claim-to-evidence audit are reported in the supplement.

\section{Belief-State JEPA}
\label{sec:architecture}

\begin{figure*}[t]
  \centering
  \begin{subfigure}[b]{0.48\textwidth}
    \centering
    \includegraphics[width=\linewidth]{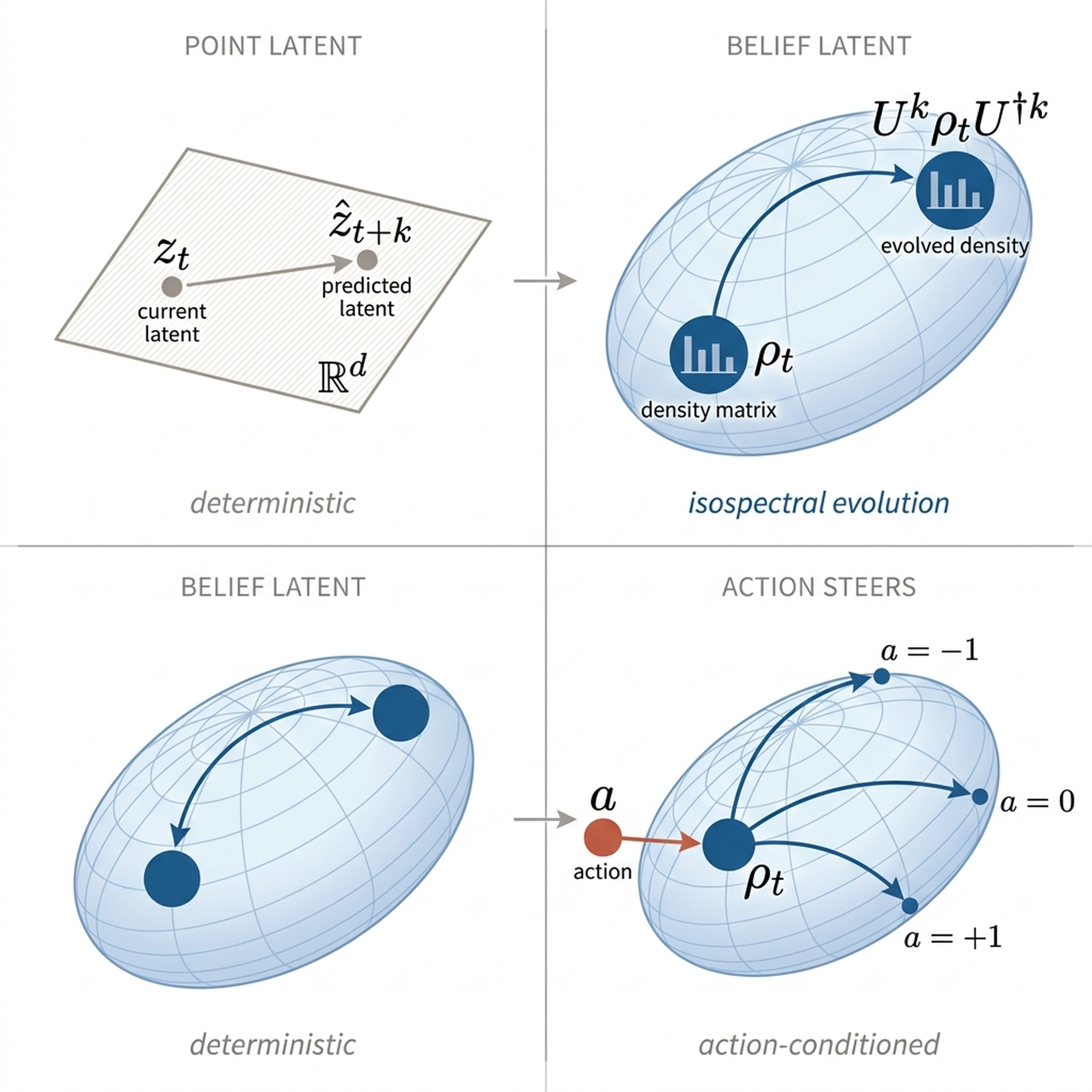}
    \caption{Concept: point vs.\ belief latent.}
    \label{fig:hero}
  \end{subfigure}
  \hfill
  \begin{subfigure}[b]{0.48\textwidth}
    \centering
    \includegraphics[width=\linewidth]{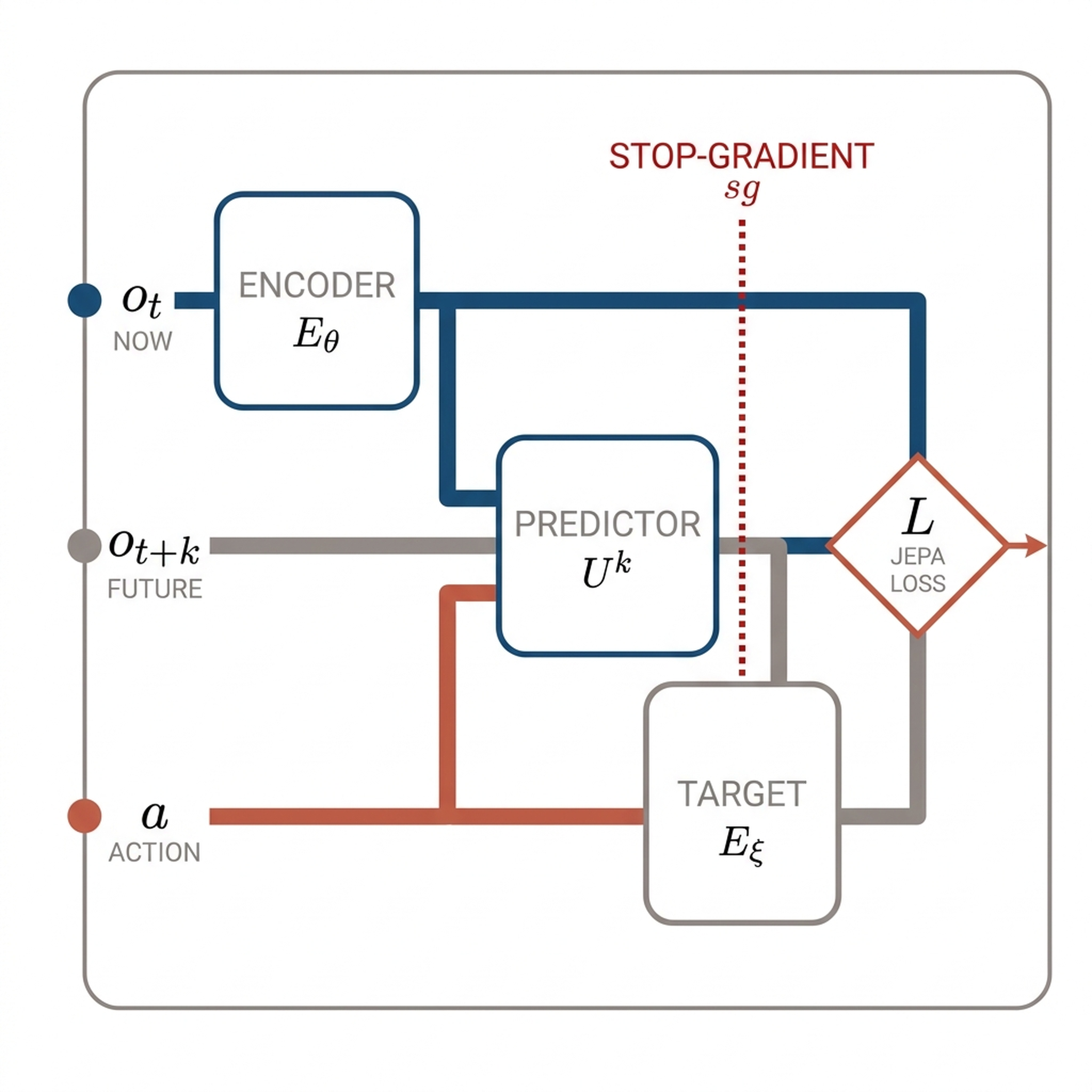}
    \caption{Architecture: JEPA scaffold with unitary predictor.}
    \label{fig:architecture}
  \end{subfigure}
  \caption{\textbf{From point latents to belief-structured imagination,
  and the architecture that realises it.}
  \textbf{(\subref{fig:hero})} A standard vector-latent JEPA predicts a
  point in representation space. UWM-JEPA instead represents the latent
  as a density matrix and rolls it forward on an isospectral orbit.
  Under partial observability this gives the predictor a structured
  latent in which uncertainty and hidden modes can be carried through
  blind rollout; actions steer the unitary trajectory through
  $H(a)=H_0+aH_1$.
  \textbf{(\subref{fig:architecture})} The online encoder maps the
  current observation history to $\rho_t$; the predictor applies
  $U^k\rho_t(U^\dagger)^k$ or an action-conditioned sequence
  $U(a_{t+k-1})\cdots U(a_t)\rho_t U(a_t)^\dagger\cdots U(a_{t+k-1})^\dagger$.
  The EMA target encoder supplies the stop-gradient future target, and
  the JEPA loss is evaluated after the system readout and projection.}
  \label{fig:overview}
\end{figure*}

UWM-JEPA modifies the \emph{predictor} of a JEPA-family world model
and leaves the surrounding training protocol unchanged. Given a
partially observed sequence $o_{\le t}$, the online encoder
$E_\theta$ produces a joint density matrix
\begin{equation}
  \rho_t \in \mathbb{C}^{\dtot \times \dtot},
  \qquad
  \rho_t=\rho_t^\dagger,\quad
  \rho_t\succeq 0,\quad
  \Tr(\rho_t)=1,
  \label{eq:density-latent}
\end{equation}
on a bipartite latent space
$\mathcal{H}_{\tot}=\mathcal{H}_{\sys}\otimes\mathcal{H}_{\env}$~\cite{nielsen2010quantum,breuer2002theory}.
All experiments use $\dsys=8$, $\denv=2$, and $\dtot=16$. The system
factor is the part read by the JEPA loss and by downstream probes; the
environment factor gives the model a register in which hidden modes
and system--environment correlations can be carried during rollout.
The practical readout is the reduced state
$\rho_t^{\sys}=\Tr_{\env}\rho_t$~\cite{nielsen2010quantum}.

The density-matrix constraints are enforced by construction. The
initial state is maximally mixed on the joint space; each observation
update sandwiches the current state by a positive system measurement
operator lifted as $K(o_t)\otimes I_{\env}$ and then renormalises to
unit trace; each prediction step is unitary conjugation. These
operations preserve Hermiticity and positive semidefiniteness up to
numerical symmetrisation, so the conditions in
\cref{eq:density-latent} are an invariant of the recurrence rather
than an unconstrained decoder output.

The prediction operator is the central architectural choice. For a
horizon $k$, UWM-JEPA rolls the latent forward by unitary conjugation,
\begin{equation}
  P_k(\rho_t)=U^k\rho_t(U^\dagger)^k,
  \qquad
  U=\exp(-iH\Delta t),\quad H=H^\dagger.
  \label{eq:predictor}
\end{equation}
Blind rollout therefore traces the unitary orbit of the joint density
matrix rather than an unconstrained vector trajectory. The target
branch follows the usual JEPA construction: an EMA encoder $E_\xi$
produces $\tilde\rho_{t+k}=E_\xi(o_{\le t+k})$ with stop-gradient on
the target parameters~\cite{grill2020bootstrap,tarvainen2017mean,caron2021emerging,chen2021exploring},
\begin{equation}
  \xi \leftarrow \tau \xi + (1-\tau)\theta,
  \qquad
  \nabla_\xi \mathcal{L}_{\mathrm{JEPA}}=0.
  \label{eq:ema-target}
\end{equation}
The loss is evaluated after reduction to the system factor and after a
small projection head $g_\phi$. Writing
$\hat h_{t+k}=g_\phi(\Tr_{\env}P_k(E_\theta(o_{\le t})))$ and
$h^+_{t+k}=g_\phi(\Tr_{\env}E_\xi(o_{\le t+k}))$, the objective is
\begin{equation}
  \mathcal{L}_{\mathrm{JEPA}}
  =
  \left\|
  \hat h_{t+k}-\mathrm{sg}\!\left[h^+_{t+k}\right]
  \right\|_F^2 .
  \label{eq:jepa-loss}
\end{equation}
The model therefore exposes two readouts. The structural guarantee
in the next section is exact on the \emph{joint} state, while the
learned objective and all downstream probes operate after partial
trace and projection.

Actions enter through an action-conditioned generator,
$H(a)=H_0+aH_1$~\cite{horowitz2022quantum,greydanus2019hamiltonian},
with rollout governed by the corresponding sequence of unitaries. The
action experiments below compare teacher-forced and counterfactual
\emph{targets} while holding the JEPA scaffold fixed, isolating the
target-construction effect from architectural changes to the predictor.

\section{Exact Non-Forgetting, Operational Memory}
\label{sec:nonforget}

The term \emph{without forgetting} has a precise scope in this paper:
the predictor channel is non-dissipative on the joint density matrix,
even though individual downstream probes may not retain their accuracy
at every horizon. For $P_k(\rho)=U^k\rho(U^\dagger)^k$, unitary
conjugation preserves the joint spectrum and therefore every spectral
functional of the joint state~\cite{nielsen2010quantum}:
\begin{equation}
  \begin{aligned}
  \mathrm{spec}(P_k(\rho))&=\mathrm{spec}(\rho),\\
  \Tr(P_k(\rho)^2)&=\Tr(\rho^2),\\
  S(P_k(\rho))&=S(\rho).
  \end{aligned}
  \label{eq:joint-invariants}
\end{equation}
These identities encode the structural reason the predictor cannot
dissipate information during blind
rollout~\cite{nielsen2010quantum,baker2022quantum}.
The implementation preserves these invariants to numerical precision
over $30$ rollout steps; the largest observed drift is below
$2.4\times 10^{-7}$ (Supplementary \cref{fig:supp-invariants}).

The qualification matters as much as the guarantee. The JEPA loss and
the probes observe $\Tr_{\env}\rho$ through a projection rather than
the full joint state, and partial trace is not unitary. Information can
move into system--environment correlations and become inaccessible to a
reduced-state linear probe while the joint state remains exactly
non-dissipative. The paper therefore reports two levels of evidence:
\cref{eq:joint-invariants} fixes the structural property of the
predictor, and the blind-rollout experiments in \cref{sec:imagine}
measure how much task-relevant information survives reduction,
projection, and repeated prediction without new observations.

The same geometry yields a useful diagnostic for what a unitary
predictor can match on the full joint space. The spectrum-mismatch
theorem in \cref{sec:theorem} states that
\begin{equation}
  \min_{U\in \mathrm{U}(\dtot)}
  \|U\rho U^\dagger-\sigma\|_F^2
  =
  \|\lambda^\downarrow(\rho)-\lambda^\downarrow(\sigma)\|_2^2.
  \label{eq:spectrum-mismatch-short}
\end{equation}
A unitary predictor can therefore match a full-state target exactly
only when context and target density matrices are isospectral. The
statement constrains the predictor's orbit geometry on the joint state
and does not transfer as a lower bound to the reduced and projected
JEPA loss in \cref{eq:jepa-loss}. In training, the online and EMA
target encoders co-adapt, so the objective can become small by learning
spectrum-compatible regions of latent space before the
reduced/projected loss is evaluated. The full theorem, proof, and a
numerical orbit visualisation appear in the supplement.

\section{Result I: Counterfactual Actions Require Counterfactual Targets}
\label{sec:actions}

An action-conditioned world model should change its imagined future
when the action sequence changes. Under a teacher-forced JEPA objective
the target encoder observes the trajectory that already contains the
action's effect, which admits an action-invariant solution in which the
predictor matches the target without using the explicit action channel.
We therefore separate two questions. Does the training target make the
action pathway identifiable? Once the pathway is identified, does the
imagined state support an action-dependent downstream readout?

\subsection{Action-Conditioned Predictor}
\label{sec:action-predictor}

The static generator $H$ is replaced by a linear action-conditioned
family~\cite{horowitz2022quantum}
\begin{equation}
H(a) = H_0 + a \cdot H_1,
\qquad U(a) = \exp(-i\, H(a)\, \Delta t),
\label{eq:action-ham}
\end{equation}
with $H_0$ and $H_1$ learnable Hermitian parameters. Discretising
$a \in \{-2, -1, 0, +1, +2\}$ allows the five matrices $U(a)$ to be
pre-computed for the action alphabet and gathered per sample per step.
For an input sequence of actions $(a_t, a_{t+1}, \dots)$ the rollout is
$\hat\rho_{t+k} = U(a_{t+k-1}) \cdots U(a_t) \, \rho_t \, U(a_t)^\dagger
\cdots U(a_{t+k-1})^\dagger$, which uses the same two batched matrix
multiplications per rollout step as the unconditioned predictor plus
an $O(|\mathcal{A}|d^3)$ matrix-exponential precompute for the
discrete action set~\cite{oh2015action,watter2015e2c}.

\subsection{Teacher Forcing Makes Actions Inert}
\label{sec:teacher-forced}

The natural training target for \cref{eq:action-ham} is the
teacher-forced JEPA objective: at each step, match the EMA target
$z_{t+k}^+ = E_\xi(x_{\le t+k})$ produced from the observed trajectory,
which carries the true action history. Under this target,
$\|H_1\| / \|H_0\| \approx 0.03$ at the end of training: the action
channel is driven to near-zero magnitude and $U(a) \approx U(0)$ for
every discrete $a$. Because the observed trajectory already contains
the action's effect and the loss does not distinguish action-sensitive
from action-invariant solutions, the predictor learns to reproduce the
observed trajectory regardless of the action. Teacher forcing alone
leaves the action pathway inert under a JEPA target.

\subsection{Counterfactual Targets Restore Action Binding}
\label{sec:counterfactual}

The fix is to replace the target by a simulator rollout under a
\emph{freshly sampled counterfactual action sequence}
$\tilde a_{t:t+k-1}$~\cite{buesing2018woulda}. Define
\begin{equation}
\tilde\rho_{t+k}^+ \;=\; E_\xi\bigl(x_{\le t} \;\Vert\;
\mathsf{Sim}(s_t, \tilde a_{t:t+k-1})\bigr),
\label{eq:cf-target}
\end{equation}
where $\mathsf{Sim}(s_t, \tilde a_{t:t+k-1})$ is the environment
simulator advanced from the latent simulator state $s_t$ under the
counterfactual actions, and $\Vert$ denotes concatenation of the
observation history. The predictor is then trained to match
$\tilde\rho_{t+k}^+$ when fed $\tilde a_{t:t+k-1}$. Sampling $\tilde a$
independently per batch makes two different action sequences map to
two different targets, so any $H_1$ that is too small incurs loss. The
construction requires simulator-state access during training and is
therefore a controlled test of counterfactual action binding rather
than a recipe for raw video streams without a counterfactual data
source.

After training with the counterfactual target,
$\|H_1\| / \|H_0\| = 1.00 \pm 0.15$ across three seeds: the two terms
in \cref{eq:action-ham} become comparable in magnitude, as expected of
a predictor that uses the action. We refer to this model below as
UWM-JEPA-CF (the suffix CF marks counterfactual-target training) and
to the matched recurrent baseline trained under the same protocol as
LSTM-JEPA-CF.

\subsection{Action Control Battery}
\label{sec:action-controls}

To confirm that the UWM-JEPA-CF predictor actually uses the action channel
(rather than achieving small loss by memorising action-averaged
dynamics), we run action-perturbation controls at test time. Each
condition reports the Hilbert--Schmidt distance from the final
predicted state to the correct counterfactual target; smaller is
better.

\begin{table}[t]
\centering
\footnotesize
\setlength{\tabcolsep}{3pt}
\caption{\textbf{Action perturbation controls on UWM-JEPA-CF.}
Mean $\pm$ std over three seeds. Correct actions give the smallest
distance among test-time perturbations of the same trained model;
wrong, shuffled, and negated actions increase it. The reversed sequence
is a mild perturbation here, so the stronger action-ordering claim rests
on the downstream indicator task. Across the same seeds,
$\|[H_0,H_1]\|/(\|H_0\|\|H_1\|)=0.68\pm0.09$.}
\label{tab:action-controls}
\begin{tabularx}{\columnwidth}{@{}Xr@{}}
\toprule
Test-time action condition & HS distance to target \\
\midrule
Correct actions      & $\mathbf{0.0082 \pm 0.0012}$ \\
Reversed sequence    & $0.0090 \pm 0.0014$ \\
Base dynamics        & $0.0096 \pm 0.0016$ \\
Wrong actions        & $0.0114 \pm 0.0021$ \\
Shuffled actions     & $0.0115 \pm 0.0020$ \\
Negated actions      & $0.0132 \pm 0.0022$ \\
\bottomrule
\end{tabularx}
\end{table}

\subsection{Hidden-Velocity Indicator Task}
\label{sec:indicator}

The counterfactual-target and perturbation results establish that
UWM-JEPA-CF uses the action channel. The indicator task tests whether
this translates into a downstream ability that is absent in the
matched LSTM-JEPA-CF configuration evaluated in the main comparison.

\paragraph{Task}
The indicator is a binary classification: predict
$y = \mathbb{1}[\text{hidden velocity at } t+K > 0]$ with $K = 5$.
The observation at time $t+K$ is masked in the input (observation
stride $3$), so the answer cannot be read off an observed frame. The
model must roll the latent $K$ steps forward \emph{under the action
sequence} and read the sign of velocity from its own imagined state.
The task therefore probes whether the action-conditioned predictor
produces a state from which the hidden continuation is linearly
decodable.

\begin{figure*}[t]
\centering
\includegraphics[width=\textwidth]{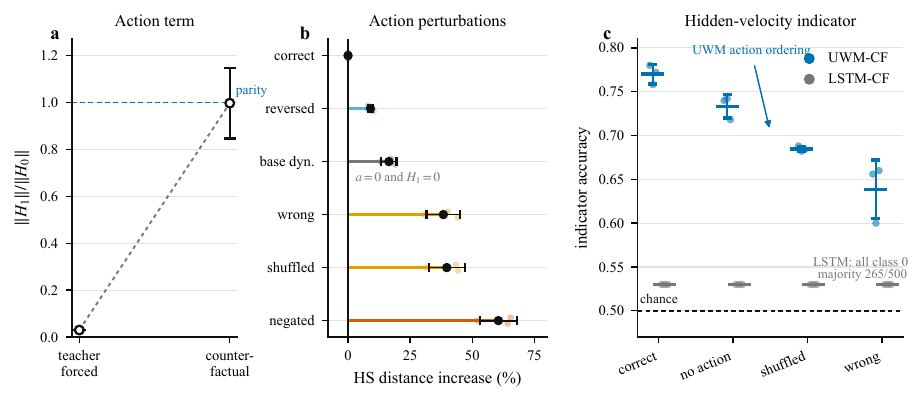}
\caption{\textbf{Action-binding evidence.} \textbf{A}: teacher-forced
JEPA makes the action Hamiltonian nearly inert, while counterfactual
targets recover an action term comparable to the base dynamics.
\textbf{B}: action perturbation controls increase Hilbert--Schmidt
distance to the counterfactual target. \textbf{C}: the hidden-velocity
indicator is solved above chance by UWM-JEPA-CF, while the matched
LSTM-JEPA-CF main-comparison configuration collapses to majority-class
predictor.}
\label{fig:action-binding}
\end{figure*}

\begin{table}[t]
\centering
\footnotesize
\setlength{\tabcolsep}{3pt}
\caption{\textbf{Hidden-velocity indicator accuracy.} Values are
mean $\pm$ std over three seeds. UWM-JEPA-CF degrades under action
perturbations; the matched LSTM-JEPA-CF configuration stays at
majority-class accuracy.}
\label{tab:indicator}
\begin{tabularx}{\columnwidth}{@{}lRR@{}}
\toprule
Condition & UWM-JEPA-CF & LSTM-JEPA-CF \\
\midrule
Correct   & $\mathbf{0.770 \pm 0.011}$ & $0.530 \pm 0.000$ \\
No-action & $0.733 \pm 0.013$ & $0.530 \pm 0.000$ \\
Shuffled  & $0.685 \pm 0.003$ & $0.530 \pm 0.000$ \\
Wrong     & $0.639 \pm 0.034$ & $0.530 \pm 0.000$ \\
\bottomrule
\end{tabularx}
\end{table}

\paragraph{Reading}
Three findings emerge (\cref{fig:action-binding}, \cref{tab:indicator}).
First, UWM-JEPA-CF solves the task above chance ($0.770$ with correct
actions) and is action-sensitive: correct exceeds wrong by $13$
percentage points, and the ordering across conditions (correct $>$
no-action $>$ shuffled $>$ wrong) matches physical intuition, since a
wrong action sequence injects the most incorrect velocity signal while
no-action injects none and biases only mildly away from the true
continuation. Second, the parameter-matched LSTM-JEPA-CF, trained on
the same counterfactual JEPA objective with the same action head, is
flat at $0.530$ across all four action conditions: all three
LSTM-JEPA-CF checkpoints predict class $0$ for all $500$ held-out
examples under every action condition, against $265$ class-$0$ and
$235$ class-$1$ ground-truth examples. Third, the scope of the claim
is fixed: under the same JEPA+counterfactual protocol with matched
parameter count, the LSTM variant fails this task while UWM-JEPA-CF
solves it above chance and with the expected action-ordering.

Additional LSTM-CF and supervised recurrent controls are reported in
\cref{sec:baseline-ablations}.

\section{Result II: Blind Rollout Tests Predictor Geometry}
\label{sec:imagine}

The spectral theorem (\cref{sec:nonforget}) preserves the joint-state
spectrum exactly but says nothing about whether useful reduced-state
information survives a long imagined trajectory. This section measures
how much of the target encoder's representation is recoverable after
$k$ blind predictor steps, and compares UWM-JEPA against a
parameter-matched LSTM-JEPA under a single protocol.

\paragraph{Protocol}
Training uses a partially observed CartPole-style spring benchmark with
a JEPA objective (online encoder, EMA target encoder, latent-space
loss, stop-gradient on the target
branch)~\cite{hafner2020dream,hafner2019learning}. All models in this
experiment are trained at horizon $k=5$, so $k=1$ and $k=3$ test
intermediate rollout coherence while $k=10$ and $k=20$ test
extrapolation beyond the trained horizon. At evaluation time we fix a
context $x_{\le t}$, obtain the predicted latent
$\hat z_{t+k} = P_k(E_\theta(x_{\le t}))$, and compare it against the
target latent $z_{t+k}^+ = E_{\xi}(x_{\le t+k})$. For each horizon
$k \in \{1, 3, 5, 10, 20\}$ a frozen linear probe maps latent to
ground-truth hidden state and reports $R^2$~\cite{alain2017understanding}.
The \emph{teacher} $R^2$ scores the probe on $z_{t+k}^+$ (the target
encoder has seen the true observation at time $t+k$); the \emph{blind}
$R^2$ scores it on $\hat z_{t+k}$ (the predictor has seen no
observations after $t$). The retention metric is
$\Delta R^2 = R^2_{\text{teacher}} - R^2_{\text{blind}}$, the
information lost relative to what was in principle recoverable at that
horizon. All values are mean $\pm$ std across three seeds.

\begin{figure*}[t]
\centering
\includegraphics[width=\textwidth]{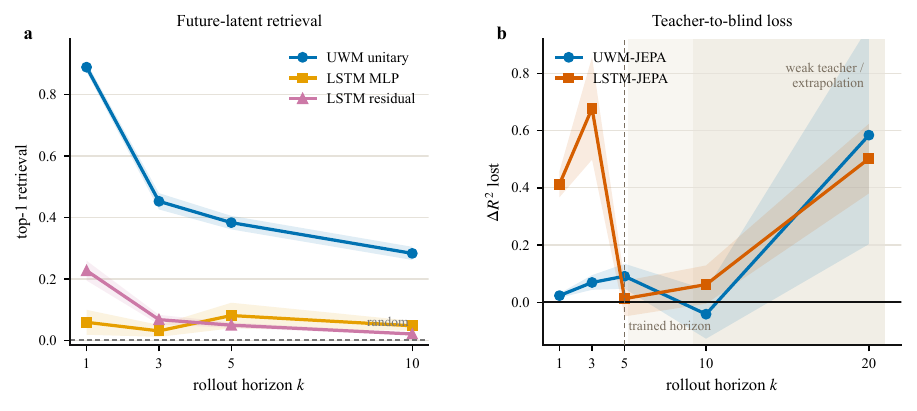}
\caption{\textbf{Blind-rollout evidence.} \textbf{A}: top-1 retrieval
accuracy for predicted future latents, comparing the UWM unitary
predictor to LSTM-JEPA encoders with plain and residual MLP predictors.
\textbf{B}: information lost under blind rollout, measured as
$\Delta R^2=R^2_{\mathrm{teacher}}-R^2_{\mathrm{blind}}$ for frozen
probes. Lower is better. The dashed vertical line marks the trained
horizon ($k=5$); horizons beyond it are extrapolations, not monotone
information-theoretic forgetting curves. The shaded region marks the
high-variance weak-teacher regime ($k\ge 10$). Together the two panels
show that the context-probe tie in \cref{fig:heldout} does not translate
into matched predictor behaviour.}
\label{fig:blind-rollout}
\end{figure*}

\begin{table*}[t]
\centering
\footnotesize
\setlength{\tabcolsep}{6pt}
\caption{\textbf{Blind-rollout information loss.}
$\Delta R^2=R^2_{\mathrm{teacher}}-R^2_{\mathrm{blind}}$, mean $\pm$
std over three seeds. Lower is better; the headline comparison is
$k=1,3$, before the trained horizon.}
\label{tab:retention}
\begin{tabular}{@{}lccccc@{}}
\toprule
Model & $k{=}1$ & $k{=}3$ & $k{=}5$ & $k{=}10$ & $k{=}20$ \\
\midrule
UWM  & $\mathbf{0.023\pm0.008}$ & $\mathbf{0.069\pm0.026}$ & $0.091\pm0.044$ & $-0.042\pm0.086$ & $0.584\pm0.381$ \\
LSTM & $0.413\pm0.046$ & $0.678\pm0.179$ & $\mathbf{0.013\pm0.062}$ & $0.062\pm0.067$ & $0.502\pm0.122$ \\
\bottomrule
\end{tabular}
\end{table*}

\paragraph{Reading}
UWM-JEPA shows strong short-horizon retention at
$k{=}1$ and $k{=}3$, remains competitive at the trained horizon
$k{=}5$, and degrades by long horizons (\cref{fig:blind-rollout},
panel B; \cref{tab:retention}). Concretely, UWM's $\Delta R^2$ is
$0.023$ at $k{=}1$, $0.069$ at $k{=}3$, and $0.091$ at $k{=}5$: blind
rollout loses less than ten points of $R^2$ relative to what the
target encoder extracts from the true observation at the same horizon.
The matched LSTM-JEPA trained on the same
objective~\cite{hochreiter1997long} has much larger $\Delta R^2$ at
$k{=}1$ and $k{=}3$ ($0.413$ and $0.678$), indicating that its
intermediate blind states are poorly aligned with the target
representation. The LSTM-JEPA re-aligns at $k{=}5$, which is exactly
the horizon used during training; this non-monotonicity is a property
of the learned predictor and probe geometry rather than a physically
monotone information curve. At $k{=}10$ the UWM mean $\Delta R^2$ is
slightly negative, within the reported seed variation, and by $k{=}20$
both models have weak teacher probes and negative blind fits. The
ratio $R^2_{\text{blind}}/R^2_{\text{teacher}}$ is therefore not
reported at $k{=}20$, since dividing two noisy numbers near zero
provides no additional signal.

Short-horizon UWM-JEPA preserves target-nearness better than the
parameter-matched LSTM-JEPA, tracks it at mid horizons, and loses the
signal at long horizons that no tested predictor handles reliably. The
retrieval panel reinforces the same separation: the UWM predictor
improves future-latent retrieval while the LSTM-JEPA models with plain
and residual MLP predictors remain well below it. The pattern is
consistent with the theorem, which constrains only the joint state,
but does not follow from it.

\section{Result III: Context Is Not the Explanation}
\label{sec:representation}

The behavioural results above admit an alternative reading in which
UWM-JEPA owes its separation to a stronger context encoder rather than
to its predictor. The held-out probe in this section tests that
reading directly. UWM-JEPA and a parameter-matched LSTM-JEPA agree
within seed noise on the same probe, which locates the empirical
separation in the predictor rather than in encoder quality.

\subsection{Held-out context probe}

\begin{figure}[t]
  \centering
  \includegraphics[width=\linewidth]{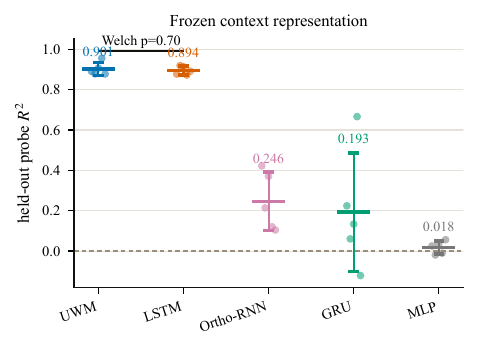}
  \caption{\textbf{Held-out linear-probe $R^2$ on hidden velocity
  (five seeds).} UWM-JEPA ($0.901 \pm 0.032$) and LSTM-JEPA
  ($0.894 \pm 0.021$) are not significantly different
  (Welch $p=0.70$). The density-matrix latent is not a win on context
  probing alone; the orthogonal, GRU, and MLP baselines fall far below.}
  \label{fig:heldout}
\end{figure}

We freeze each trained encoder, take the latent produced on a short
context window, and fit a linear probe to predict the hidden velocity
of the next step~\cite{alain2017understanding}. Training and evaluation
splits are disjoint across trajectories and the probe is linear, so the
metric reflects information in the encoder's representation rather than
downstream nonlinear capacity. Figure~\ref{fig:heldout} reports
five-seed results.

The two top entries in \cref{fig:heldout} sit within error of each
other: UWM-JEPA at $0.901 \pm 0.032$, LSTM-JEPA at $0.894 \pm 0.021$,
with Welch $p=0.70$. A well-regularised recurrent encoder already
recovers nearly all of the probe-accessible information about
hidden velocity in this domain, so the density-matrix latent confers
no context-probing advantage. The three non-recurrent baselines in
the same figure (orthogonal~\cite{arjovsky2016unitary,wisdom2016full},
GRU~\cite{cho2014learning}, and MLP at $0.246 \pm 0.144$,
$0.193 \pm 0.293$, and $0.018 \pm 0.032$) confirm that the probe
discriminates: it places the two strong recurrent and density-matrix
encoders together at the top and the weaker baselines well below them.
The GRU variance is large, so its mean is not interpreted sharply; the
relevant point is that the UWM--LSTM tie at the top reflects shared
high probe quality, not a ceiling on the probe itself.

\paragraph{Consequence}
Since the context probe ties, the remaining experiments evaluate the
predictor rather than the encoder. We test whether predicted future
latents retrieve the right target latent, whether frozen probes retain
hidden-state information after blind rollout, and whether counterfactual
actions move the latent in the expected direction. Collapse diagnostics
and anti-collapse sweeps are reported in the supplement; they confirm
that the UWM-JEPA runs used here remain non-collapsed under the tested
regularisation protocols but use different regulariser variants and
checkpoint protocols, so they are not directly comparable to the
five-seed context-probe numbers in \cref{fig:heldout}.

\section{Discussion and Scope}
\label{sec:limitations}

UWM-JEPA is a JEPA-family world model whose latent is a density matrix
and whose predictor is unitary on a system--environment space. This
construction yields an exact non-dissipation property on the joint
latent and, in the tested partially observable settings, improves
intermediate-horizon blind rollout and counterfactual
action-conditioned imagination. The architecture is not a universally
stronger encoder, a general anti-collapse mechanism, or a competitive
planner.

The phrase \emph{belief state} is used here in an operational sense:
the latent is a density-matrix representation capable of carrying
uncertainty and system--environment correlations through rollout.
Calibrated posterior structure over the true hidden state would
require a separate posterior-alignment experiment and is outside the
present claims.

The headline negative result is the context-probe tie. UWM-JEPA and
LSTM-JEPA encode the observed prefix equally well on the held-out
hidden-velocity probe, so the empirical advantage appears only when
the latent is rolled forward without new observations or when actions
are varied counterfactually. This separation organises the experiments
around encoder quality, blind rollout, and action binding rather than
around a single aggregate score.

The non-forgetting theorem applies to the joint density matrix.
Unitary conjugation preserves the spectrum, purity, and entropy of the
\emph{joint} state, but the downstream loss and probes see only the
reduced and projected system state. Reduced-state information can
move into correlations and become inaccessible to a linear probe even
when the joint evolution is exactly reversible, so the operational
memory curves in \cref{fig:blind-rollout} complement rather than
duplicate the theorem; they measure what remains useful after the
readout used by the actual JEPA objective. The JEPA predictors in
that experiment are trained at horizon $k=5$, so $k=1$ and $k=3$
probe intermediate rollout coherence and $k>5$ probes extrapolation
rather than a monotone information-theoretic forgetting curve.

The action results establish counterfactual sensitivity rather than
broad planning competence. Teacher-forced action JEPA leaves the action
Hamiltonian nearly inert, whereas counterfactual targets recover an
action term comparable to the base dynamics and yield an above-chance
hidden-velocity indicator with the expected degradation under wrong,
shuffled, and no-action controls. The construction is evidence for
action-conditioned imagination in this testbed, not for long-horizon
planning, and not for alternative recurrent or Transformer objectives.
The counterfactual training target also requires simulator access to
construct alternate futures, which limits portability to settings where
only passive videos or observation streams are available.

Matched-baseline language refers to trainable parameter count rather
than the number of coordinates in the latent state. A $16\times16$
density matrix carries more real degrees of freedom than a length-$16$
vector hidden state, even when the trainable encoder and predictor
budgets are matched, which makes the context-probe tie in
\cref{fig:heldout} a central control: it shows that the main
comparison is not explained by a better frozen context representation.
The supplementary baseline ablations further separate target-encoder
decodability from blind-rollout performance.

The present experiments use structured partially observable oscillator
and CartPole-style settings, and the present results do not extend to
hidden Markov chains, ring-walker environments, or unstructured visual
prediction. The current comparison set covers LSTM-JEPA models with
plain and residual MLP predictors, GRU, orthogonal recurrent, and MLP
baselines, with the recurrent families grounded in
GRU~\cite{cho2014learning}, orthogonal/unitary
RNN~\cite{arjovsky2016unitary,wisdom2016full}, and
LSTM~\cite{hochreiter1997long} architectures. Learned Lie-algebra
predictors~\cite{greydanus2019hamiltonian}, neural ODE
predictors~\cite{chen2018neural}, state-space
models~\cite{gu2022efficiently,gu2023mamba}, and Transformer
predictors~\cite{vaswani2017attention} remain natural future
baselines. Within that scope, the contribution is a controlled
demonstration that belief-state structure can separate context
representation from rollout-time memory in a JEPA world model.

Seed counts are asymmetric by design. The frozen context probe is
reported over five seeds because it is the key negative control for
encoder quality; blind-rollout and action-conditioned experiments are
reported over three seeds because they require trained predictor or
counterfactual-action checkpoints. Strong claims are confined to
low-variance regions and seed counts appear in every relevant table
caption.

{
\small
\bibliography{references}
}

\vspace{8pt}
\noindent\rule{\columnwidth}{0.4pt}
\vspace{4pt}

\section*{Data and Code Availability}
All code, data, and figure-generation scripts are available at \url{https://github.com/santoshkumarradha/uwm-jepa}. The evidence figures in the paper are regenerated from the experiment outputs bundled in the \texttt{experiments/data/} folder of that repository.

\clearpage
\onecolumn
\appendix

\setcounter{figure}{0}
\setcounter{table}{0}
\setcounter{equation}{0}
\renewcommand{\thefigure}{S\arabic{figure}}
\renewcommand{\thetable}{S\arabic{table}}
\renewcommand{\theequation}{S\arabic{equation}}
\renewcommand{\thesection}{S\arabic{section}}
\renewcommand{\thesubsection}{S\arabic{section}.\arabic{subsection}}

\begin{center}
  \vspace{20pt}
  {\Large\bfseries Supplementary Information}\\[10pt]
  {\large UWM-JEPA: Predictive World Models That Imagine in Belief Space}\\[6pt]
  {Santosh Kumar Radha and Oktay Goktas}
  \vspace{12pt}
\end{center}

\noindent\rule{\textwidth}{0.4pt}
\vspace{10pt}

\section{Spectrum-Mismatch Lower Bound}\label{sec:theorem}

A JEPA whose latent predictor is a unitary map on density matrices faces
a structural obstruction that depends only on spectra. The closed-form
lower bound below characterises the Hilbert--Schmidt training loss of
such a predictor: the bound vanishes if and only if the context and
target density matrices have identical spectra. The main text invokes
this result as a structural diagnostic for the full joint-state unitary
predictor. The practical training objective is reduced and projected,
so the theorem does not transfer as an exact lower bound on the loss
optimised in \cref{eq:jepa-loss}.

\subsection{Setup}\label{sec:theorem-setup}

Let $\mathcal{H} \cong \mathbb{C}^{d}$ with $d = \dtot$ be the joint
system--environment Hilbert space of the architecture introduced in
\cref{sec:architecture}~\cite{nielsen2010quantum,breuer2002theory}, and let $\mathrm{U}(d)$ denote the group of
$d \times d$ unitary matrices. For a Hermitian matrix $A \in \mathbb{C}^{d
\times d}$, write $\lambda^{\downarrow}(A) \in \mathbb{R}^{d}$ for its
eigenvalue vector arranged in non-increasing order. The
Hilbert--Schmidt (Frobenius) norm and inner product are
$\langle A, B \rangle_{F} = \Tr(A^{\dagger} B)$ and
$\lVert A \rVert_{F} = \sqrt{\Tr(A^{\dagger} A)}$.
A JEPA training pair $(\rho_{t}, \rho_{t+k}^{+})$ consists of a context
density matrix $\rho_{t}$, produced by the online encoder from observations
$x_{\leq t}$, and a target density matrix $\rho_{t+k}^{+}$, produced by the
EMA target encoder from observations $x_{\leq t+k}$. The predictor is a
unitary $U^{k} \in \mathrm{U}(d)$ obtained by $k$-fold application of the
one-step unitary in \cref{eq:predictor}, and the full-state matching loss is
\begin{equation}\label{eq:jepa-loss-main}
  \mathcal{L}(U^{k}; \rho_{t}, \rho_{t+k}^{+})
  \;=\;
  \bigl\lVert U^{k} \rho_{t}\, (U^{k})^{\dagger}
              - \rho_{t+k}^{+} \bigr\rVert_{F}^{2} .
\end{equation}

\subsection{Main theorem}\label{sec:theorem-main}

\begin{theorem}[Spectrum-mismatch lower bound]\label{thm:spectrum-mm}
Let $\rho, \sigma \in \mathbb{C}^{d \times d}$ be Hermitian, with sorted
eigenvalue vectors $\lambda^{\downarrow}(\rho), \lambda^{\downarrow}(\sigma)
\in \mathbb{R}^{d}$. Then
\begin{equation}\label{eq:spectrum-mm-main}
  \min_{U \in \mathrm{U}(d)}
  \bigl\lVert U \rho\, U^{\dagger} - \sigma \bigr\rVert_{F}^{2}
  \;=\;
  \bigl\lVert \lambda^{\downarrow}(\rho)
             - \lambda^{\downarrow}(\sigma) \bigr\rVert_{2}^{2},
\end{equation}
and the minimum is attained at any $U^{\star} \in \mathrm{U}(d)$ that maps a
sorted eigenbasis of $\rho$ to a sorted eigenbasis of $\sigma$ in matching
order.
\end{theorem}

\paragraph{Proof} Expand the Frobenius norm as
$\Tr(\rho^{2}) + \Tr(\sigma^{2}) - 2\,\Tr(U \rho U^{\dagger} \sigma)$. The
first two terms are unitary-invariant; minimising the whole is equivalent to
maximising $\Tr(U \rho U^{\dagger} \sigma)$, for which von~Neumann's trace
inequality gives
$\max_{U}\Tr(U\rho U^{\dagger}\sigma) = \sum_{i}\lambda_{i}^{\downarrow}(\rho)\,
\lambda_{i}^{\downarrow}(\sigma)$, with equality when $U$ aligns the sorted
eigenbases. Substituting and collecting terms yields the squared $\ell_{2}$
distance between sorted eigenvalue vectors. Thus the result follows from
von~Neumann's trace inequality on the unitary orbit of~$\rho$ under the
Frobenius objective. It is closely related to Hoffman--Wielandt
eigenvalue-stability bounds~\cite{hoffman1953variation}, but the
optimisation used here is the unitary-orbit trace maximisation.

\subsection{Corollaries}\label{sec:theorem-corollaries}

Two consequences specialise the theorem to JEPA training.

\begin{corollary}[Irreducible JEPA error]\label{cor:irreducible}
For every unitary predictor $U^{k} \in \mathrm{U}(d)$ and every context/target
pair $(\rho_{t}, \rho_{t+k}^{+})$,
\begin{equation}\label{eq:cor-irreducible}
  \mathcal{L}(U^{k}; \rho_{t}, \rho_{t+k}^{+})
  \;\geq\;
  \bigl\lVert \lambda^{\downarrow}(\rho_{t})
             - \lambda^{\downarrow}(\rho_{t+k}^{+}) \bigr\rVert_{2}^{2}.
\end{equation}
\end{corollary}

The right-hand side of~\eqref{eq:cor-irreducible} depends only on spectra and
is independent of $U^{k}$. It therefore cannot be reduced by any training
procedure that optimises only over $\mathrm{U}(d)$. We call this quantity the
\emph{irreducible JEPA error} of the pair $(\rho_{t}, \rho_{t+k}^{+})$.

\begin{corollary}[Orbit-projected target eliminates the floor]%
\label{cor:orbit}
Let $\rho_{t+k}^{+} = V_{+} \Lambda_{+} V_{+}^{\dagger}$ be a spectral
decomposition and let $\Lambda_{t}$ contain the sorted eigenvalues of
$\rho_{t}$. The orbit-projected target
\begin{equation}\label{eq:orbit-proj-main}
  \widetilde{\rho}_{t+k}^{\,+}
  \;:=\;
  V_{+}\, \Lambda_{t}\, V_{+}^{\dagger}
\end{equation}
has $\lambda^{\downarrow}(\widetilde{\rho}_{t+k}^{\,+}) =
\lambda^{\downarrow}(\rho_{t})$, so
$\min_{U} \lVert U\rho_{t} U^{\dagger} - \widetilde{\rho}_{t+k}^{\,+}
\rVert_{F}^{2} = 0$.
\end{corollary}

Orbit projection is the minimal full-state target rewrite that makes
the unitary matching problem reducible. Eigenvectors, which carry the
direction of evolution the predictor must learn, are preserved; only
the spectrum is replaced. The corollary is a mathematical diagnostic
and does not claim that the practical reduced/projected objective
implements orbit projection.

\subsection{Physical meaning}\label{sec:theorem-physical}

Unitary conjugation is an isometry of Hermitian matrices in the
Hilbert--Schmidt inner product, so the image
$U^{k}\rho_{t}(U^{k})^{\dagger}$ lies on the \emph{unitary orbit} of
$\rho_{t}$, a submanifold of density-matrix space parametrised entirely
by the spectrum $\lambda^{\downarrow}(\rho_{t})$. A generic target
density matrix produced by encoding a future observation has a
different spectrum, since the entropy of a realised trajectory is not
the entropy of the uncertainty over trajectories, and therefore lies
on a different orbit. \Cref{thm:spectrum-mm} quantifies the gap: the
closest the predictor's output can get to such a target, over all
unitaries, is the squared $\ell_{2}$ distance between sorted spectra.
Training over $\mathrm{U}(d)$ cannot reduce that floor.

The corollaries describe full-state geometry. They imply nothing about
the spectra of reduced system states or about retention of task
variables by a downstream probe; those empirical questions are
evaluated in the blind-rollout and action-binding experiments.

\subsection{Numerical invariant check}\label{sec:theorem-invariants}

\begin{figure}[t]
  \centering
  \includegraphics[width=0.72\textwidth]{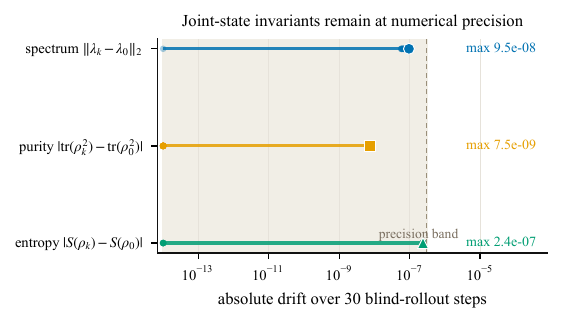}
  \caption{\textbf{Joint-state invariant check.} A trained UWM-JEPA
  predictor preserves spectrum, purity, and entropy under blind rollout
  to floating-point precision. The largest observed drift over
  $30$ steps is below $2.4\times 10^{-7}$, verifying the implemented
  unitary predictor rather than any downstream performance metric.}
  \label{fig:supp-invariants}
\end{figure}

\section{Training Stability and Collapse}
\label{sec:training}

JEPA-family objectives are non-contrastive: the encoder and predictor
are trained to match a target network's latent output, with no
explicit repulsion between unrelated samples. The known failure mode
is collapse, in which the encoder minimises the loss by mapping every
input to the same point or to a low-rank subspace. A new JEPA
predictor therefore needs evidence that it does not inherit or induce
collapse. This section (i) explains why training loss alone is not a
valid collapse diagnostic for the regularisers used in this
literature, (ii) reports a seed-level collapse phase diagram on a
held-out probe task, and (iii) runs a three-seed $\times$ six-variant
sweep of anti-collapse protocols on UWM-JEPA.

\paragraph{Training loss is not a collapse metric}
VICReg~\cite{bardes2022vicreg}, Barlow Twins~\cite{zbontar2021barlow},
and log-determinant regularisers are hinge-like or barrier-like by
construction. At the optimum the variance, cross-correlation, or
spectral terms sit near their target boundary, so the total training
loss plateaus close to a fixed non-zero value regardless of whether
the encoder has collapsed, and collapsed and healthy encoders can
produce training curves that look nearly identical. Following
standard practice in the JEPA
literature~\cite{assran2023ijepa,assran2024vjepa,caron2021emerging,chen2021exploring,grill2020bootstrap},
collapse is diagnosed downstream: we freeze the encoder after
training, fit a linear probe on a small supervised split, and report
the probe's coefficient of determination on a held-out split. A
non-trivial held-out $R^2$ is direct evidence that the latent space
retains task-relevant structure; a near-zero $R^2$ is direct evidence
of representational collapse regardless of the training loss.

\begin{figure}[t]
  \centering
  \includegraphics[width=\linewidth]{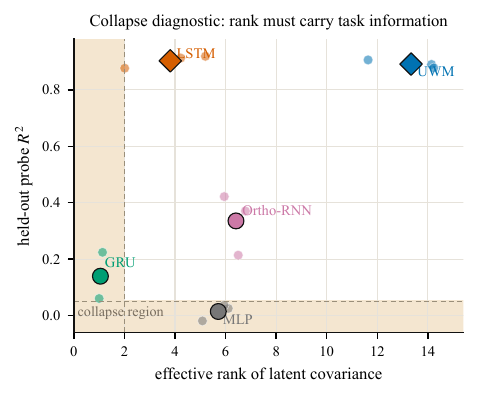}
  \caption{\textbf{Collapse diagnostic on held-out data.}
  Seed-level points compare covariance effective rank against
  linear-probe $R^2$ on hidden velocity. The shaded failure region is
  the conservative collapse rule used in this experiment
  ($R^2<0.05$ or effective rank $<2$)~\cite{roy2007effective}. UWM-JEPA and LSTM-JEPA both
  occupy the non-collapsed, high-probe region; the probe, not the
  training loss, is the primary collapse criterion.}
  \label{fig:dashboard}
\end{figure}

\paragraph{Held-out collapse diagnostic}
Figure~\ref{fig:dashboard} plots the two collapse diagnostics that are
comparable across latent parameterisations: held-out linear-probe $R^2$
and covariance effective rank~\cite{alain2017understanding,roy2007effective}. Each small point is one training seed;
the large marker is the model mean. The shaded region marks the
conservative failure rule used in the experiment ($R^2<0.05$ or
effective rank $<2$). UWM-JEPA and LSTM-JEPA both sit in the
high-probe, non-collapsed region; GRU fails primarily by low rank and
MLP fails by near-zero probe performance despite non-trivial covariance
rank. Pairwise latent distance was computed as an auxiliary diagnostic
but is not plotted because its absolute scale is not comparable between
density-matrix and vector latents.

\paragraph{Anti-collapse protocols}
To test whether UWM-JEPA's stability is an artefact of one particular
regulariser, we trained six variants of the UWM-JEPA objective across
three seeds each and evaluated each with the same held-out linear
probe. The variants are: \texttt{none} (stop-gradient target only),
\texttt{contrastive} (InfoNCE-style negatives~\cite{oord2018cpc}), \texttt{ema\_only}
(EMA target network, no explicit regulariser), \texttt{ema\_vicreg}
(EMA plus VICReg), \texttt{ema\_barlow} (EMA plus Barlow Twins), and
\texttt{ema\_logdet} (EMA plus log-determinant penalty).

\begin{table}[t]
  \centering
  \caption{\textbf{Anti-collapse sweep on UWM-JEPA (three seeds per
  variant).} Held-out linear-probe $R^2$ on hidden velocity, mean
  $\pm$ standard deviation. All six variants cross the conservative
  probe threshold on every seed; none of the runs collapsed.}
  \label{tab:multiseed}
  \begin{tabular}{lcc}
    \toprule
    Protocol & Probe $R^2$ mean & Probe $R^2$ std \\
    \midrule
    \texttt{none}          & 0.975 & 0.003 \\
    \texttt{contrastive}   & 0.921 & 0.060 \\
    \texttt{ema\_only}     & 0.975 & 0.003 \\
    \texttt{ema\_vicreg}   & 0.979 & 0.001 \\
    \texttt{ema\_barlow}   & 0.915 & 0.028 \\
    \texttt{ema\_logdet}   & 0.969 & 0.008 \\
    \bottomrule
  \end{tabular}
\end{table}

\paragraph{Scope of the claim}
Every variant in Table~\ref{tab:multiseed} crosses the conservative
probe threshold on every seed, and no run collapsed. The experiment
therefore supports the narrow statement that \emph{UWM-JEPA remained
non-collapsed under the tested protocols in this setting}. Unitarity
does not prevent collapse in general: the theorem in
\cref{sec:theorem} preserves the spectrum of the joint
system--environment state under blind rollout but does not prevent
the encoder from mapping its input to a degenerate joint state in the
first place. Across the regularisers the community typically uses to
rule collapse out, the UWM-JEPA predictor does not re-introduce it.

\section{Additional Baseline Ablations}
\label{sec:baseline-ablations}

The main action-binding result compares UWM-JEPA-CF to a
parameter-matched LSTM-JEPA-CF configuration trained with the same
counterfactual target. We also ran LSTM-CF ablations that vary
recurrent state size and learning rate while leaving the
counterfactual-JEPA target and indicator evaluation unchanged.

\begin{table}[h]
\centering
\small
\caption{\textbf{LSTM-CF tuning confirmation at the original 5k-step
budget.} Values are mean $\pm$ std over three seeds. Target accuracy
reports the linear indicator readout on target-encoder latents; correct,
wrong, no-action, and shuffled report the same readout after blind
counterfactual rollout. Larger recurrent states improve target-latent
decodability in some configurations, but do not yield a stable
action-sensitive blind rollout.}
\label{tab:lstm-cf-confirm}
\begin{tabular}{lccccc}
\toprule
Configuration & Target & Correct & Wrong & No-action & Shuffled \\
\midrule
$h{=}12$, lr $3{\times}10^{-4}$ &
$0.645\pm0.220$ & $0.531\pm0.001$ & $0.531\pm0.001$ &
$0.530\pm0.000$ & $0.531\pm0.001$ \\
$h{=}32$, lr $1{\times}10^{-4}$ &
$0.622\pm0.087$ & $0.527\pm0.006$ & $0.525\pm0.009$ &
$0.525\pm0.009$ & $0.533\pm0.005$ \\
$h{=}63$, lr $1{\times}10^{-3}$ &
$0.867\pm0.071$ & $0.479\pm0.079$ & $0.511\pm0.041$ &
$0.504\pm0.045$ & $0.487\pm0.070$ \\
$h{=}255$, lr $1{\times}10^{-3}$ &
$0.568\pm0.010$ & $0.470\pm0.000$ & $0.470\pm0.000$ &
$0.470\pm0.000$ & $0.470\pm0.000$ \\
\bottomrule
\end{tabular}
\end{table}

The $h{=}63$ row is the clearest diagnostic. The target latents support
a high-accuracy linear indicator readout, but the blind rollout does
not preserve that readout under the correct action sequence. The
failure is therefore not simply absent target-state information in the
recurrent encoder. It appears when the counterfactual-JEPA predictor is
asked to carry that information forward without new observations.

\begin{table}[h]
\centering
\small
\caption{\textbf{Directly supervised recurrent positive control.}
Values are mean $\pm$ std over three seeds. The model receives the same
context observations and action sequence, but is trained directly for
the hidden-velocity indicator rather than through a JEPA latent target.}
\label{tab:lstm-supervised-control}
\begin{tabular}{lcc}
\toprule
Configuration & Correct actions & Wrong actions \\
\midrule
$h{=}12$  & $0.960\pm0.000$ & $0.867\pm0.017$ \\
$h{=}63$  & $0.955\pm0.015$ & $0.839\pm0.001$ \\
$h{=}255$ & $0.944\pm0.002$ & $0.848\pm0.011$ \\
\bottomrule
\end{tabular}
\end{table}

The supervised control shows that action-conditioned recurrent models
can solve the indicator when optimized directly for it. Together,
\cref{tab:lstm-cf-confirm,tab:lstm-supervised-control} locate the
main-comparison failure in the tested counterfactual-JEPA latent
rollout, not in the availability of action inputs or in recurrent
capacity alone.

We also evaluated a full-state linear probe that predicts both
continuous position and hidden velocity. Across the three UWM-JEPA-CF
seeds, blind-rollout full-state mean $R^2$ is $0.291\pm0.009$ and blind
velocity-sign accuracy is $0.734\pm0.007$. In the LSTM-CF confirmation
runs, target encoders can be decodable while blind rollout remains near
majority-class behavior; for example, the $h{=}63$ confirmation gives
target full-state mean $R^2=0.845\pm0.125$ but blind velocity-sign
accuracy $0.538\pm0.007$. This broader probe supports the same
interpretation as the binary indicator: the separation appears after
blind rollout, not simply at the target-encoder representation.

\section{Claim-to-Evidence Audit}
\label{sec:claim_map}

\Cref{tab:claim_map} records the operational meaning of each major
claim in the manuscript, the evidence supporting it, and the
corresponding limitation, making the claim boundaries explicit for
audit.

\begin{table*}[t]
\centering
\small
\caption{\textbf{Claim-to-evidence map for UWM-JEPA.} Each row binds a
manuscript claim to an operational definition, the specific evidence
backing it, and an explicit limitation.}
\label{tab:claim_map}
\begin{tabular}{p{2.4cm} p{4.2cm} p{4.4cm} p{4.4cm}}
\toprule
\textbf{Title phrase} & \textbf{Operational definition} &
\textbf{Evidence} & \textbf{Limitation} \\
\midrule
UWM-JEPA &
JEPA-family latent-target training with a density-matrix representation
and a unitary predictor (encoder, EMA target, latent loss,
stop-gradient, frozen probes). &
Architecture \& loss specification (\cref{sec:architecture},
\cref{sec:training}); training dynamics in the collapse dashboard. &
Not equivalent to V-JEPA; we claim JEPA-family membership, not
method parity with any particular visual JEPA implementation. \\
\addlinespace
Imagine &
Blind, action-free latent rollout from a single observed prefix:
predict $z_{t+k}$ without access to $o_{t+1}, \ldots, o_{t+k}$. &
Retrieval-at-horizon and ${\Delta}R^2$ retention against matched
LSTM-JEPA encoders with plain and residual MLP predictors. &
Predictors trained at horizon $k{=}5$; $k{<}5$ probes intermediate
rollout coherence and $k{>}5$ probes extrapolation. Other predictor
families are not evaluated. \\
\addlinespace
Without forgetting &
(i) Exact preservation of joint-state spectral invariants under blind
rollout (theorem); (ii) empirically measurable reduced-state retention. &
Invariants numerical check ($< 10^{-7}$ drift over 30 steps); held-out
probe; $\Delta R^2$ retention at $k{=}1{-}5$. &
Reduced-state retention degrades on longer horizons; the exact
guarantee does not transfer undiminished to the probed subspace. \\
\addlinespace
World model &
Action-conditioned counterfactual prediction: swapping actions changes
predicted latent trajectories in a controlled way. &
Counterfactual H1/H0 ratio ($1.00 \pm 0.15$), nonzero commutator,
wrong-action / shuffled-action / no-action controls; hidden-velocity
indicator task; LSTM-CF tuning and supervised positive-control
ablations. &
Planning micro-task is weak (3--5$\times$ over chance); no
competitive-planner claim; counterfactual targets require simulator
state access. \\
\addlinespace
Not universal &
UWM-JEPA is presented as a structured belief-state testbed, not as a
general-purpose sequence model. &
Held-out probe tie with LSTM-JEPA ($0.901$ vs $0.894$); explicit scope
statement. &
Oscillator and CartPole-style settings only; no JEPA-native HMM or
ring-walker claim. \\
\bottomrule
\end{tabular}
\end{table*}

\paragraph{Claim ladder}
The claim hierarchy has eight rungs. (1)~UWM-JEPA is a JEPA-family
model (encoder, EMA target, latent loss, stop-gradient, frozen
probes). (2)~Its unitary predictor preserves joint-state spectral
invariants under blind rollout (theorem and numerical check). (3)~The
exact guarantee does not transfer to useful reduced-state memory;
long-horizon reduced retention is limited. (4)~On context
representation, UWM ties LSTM-JEPA ($0.901$ vs $0.894$). (5)~Under
blind rollout, UWM preserves target-nearness better than the tested
LSTM-JEPA MLP predictors at intermediate horizons. (6)~Teacher-forced
action JEPA ignores the action pathway
($\|H_1\|/\|H_0\|{\approx}0.03$). (7)~Counterfactual / internal-action
targets induce action-sensitive dynamics
($\|H_1\|/\|H_0\|{=}1.00\pm0.15$ with controls). (8)~UWM-JEPA is a
structured testbed for belief-state world models within the studied
setting.

\end{document}